\documentclass[a4paper,abstract]{scrartcl}

\usepackage{graphicx}
\usepackage{microtype}
\usepackage{paralist}
\usepackage[comma,round]{natbib}
\usepackage[hidelinks]{hyperref}
\usepackage[utf8]{inputenc}

\usepackage{times}

\deffootnote[1em]{1em}{1em}
{\textsuperscript{\thefootnotemark}}

\typearea{12}

\begin{document}

\author{Georg Rehm\thanks{Georg Rehm, DFKI GmbH, Alt-Moabit 91c, 10559 Berlin,
    Germany -- \href{mailto:georg.rehm@dfki.de}{georg.rehm@dfki.de}}}

\date{}

\title{Observations on Annotations\thanks{To be published in:
    \emph{Annotations in Scholarly Editions and Research: Functions,
      Differentiation, Systematization} (2020), Julia Nantke and Frederik
    Schlupkothen (editors). De Gruyter. In print.}}

\maketitle

\begin{abstract}
  The annotation of textual information is a fundamental activity in
  Linguistics and Computational Linguistics. This article presents various
  observations on annotations. It approaches the topic from several angles
  including Hypertext, Computational Linguistics and Language Technology,
  Artificial Intelligence and Open Science. Annotations can be examined along
  different dimensions. In terms of complexity, they can range from trivial to
  highly sophisticated, in terms of maturity from experimental to
  standardised. Annotations can be annotated themselves using more abstract
  annotations. Primary research data such as, e.\,g., text documents can be
  annotated on different layers concurrently, which are independent but can be
  exploited using multi-layer querying. Standards guarantee interoperability
  and reusability of data sets. The chapter concludes with four final
  observations, formulated as research questions or rather provocative remarks
  on the current state of annotation research.
\end{abstract}

\section{Introduction}
\label{sec:introduction}

The annotation of textual information is one of the most fundamental
activities in Linguistics and Computational Linguistics including neighbouring
fields such as, among others, Literary Studies, Library Science and Digital
Humanities \citep{ide2017,rehm2020e}. Horizontally, data annotation plays an
increasingly important role in Open Science, in the development of NLP/NLU
prototypes (Natural Language Processing/Understanding), more application- and
solution-oriented Language Technologies (LT) and systems based on neural
technologies in the area of Artificial Intelligence (AI).

This article reflects on more than two decades of research in the wider area
of annotation including multi-layer annotations \citep{witt2007b,witt2007},
the modelling of linguistic data structures
\citep{woerner2006,rehm2007d,witt2009} including hypertext and web genres
\citep{rehm2002,rehm2007,rehm2007b}, the production and distribution of
annotated corpora \citep{rehm2014c,rehm2016c,elg2020} and the use of metadata,
annotation schemes and markup languages
\citep{rehm2008d,rehm2008e,rehm2009,rehm2010}. After an initial approximation
of a definition (Section~\ref{sec:definition}), the chapter provides lessons
learned, future research directions as well as observations on the scientific
and technical process of annotating textual data from several angles including
Hypertext, Markup and the World Wide Web (Section~\ref{sec:www}),
Computational Linguistics (Section~\ref{sec:compling}), Artificial
Intelligence (Section~\ref{sec:ai}), Language Technology
(Section~\ref{sec:langtech}) and Open Science
(Section~\ref{sec:openscience}). The article concludes with an overview of the
main conceptual dimensions involved in the annotation of textual information
(Section~\ref{sec:dimensions}) and a summary (Section~\ref{sec:conclusions}).

\section{Definition}
\label{sec:definition}

Definitions of the term ``annotation'' typically focus on either procedural
(i.\,e., process-related), technical (i.\,e., markup-related) or conceptual
(i.\,e., semantics-related) aspects, sometimes also combinations of the
different layers \citep{goecke2010,ide2017}. The notion we follow in this
article is loosely based on the concept of Annotation Graphs \citep{bird2001},
which can be used to represent an unlimited number of annotation layers, while
links between the text and annotations can be established in an unrestricted
way \citep{witt2007,ide2007}. Specifically, we view annotations as
\emph{secondary research data} added to \emph{primary research
  data}. Annotations are, therefore, part of the metadata that also include
general information on the primary data (author/creator, modality, creation
date etc.). \emph{Linguistic} annotations, then, cover ``any descriptive or
analytic notations applied to raw language data. The basic data may be in the
form of [\dots] audio, video and/or physiological recordings [\dots] or it may
be textual. The added notations may include transcriptions of all sorts (from
phonetic features to discourse structures), part-of-speech and sense tagging,
syntactic analysis, ‘named entity’ identification, co-reference annotation,
and so on.'' \citep{bird2001}. The procedure of annotating data can include,
among several other variants, highlighting and labelling specific segments,
commenting upon certain aspects, and selecting as well as inserting markup
elements (tags) into a text document. The design of a concrete annotation
scheme typically follows at least two consecutive phases: based on linguistic
theory or insights, an annotation model is created \citep{pustejovsky2017a}
for which, then, a technical representation is developed
\citep{ide2017b}. \cite{finlayson2017} provide an overview of the processes
and tools involved in the creation of annotations.

\section{Hypertext, Markup and the World Wide Web}
\label{sec:www}

Annotations have always been an integral concept of hypertext \citep{nelson87}
itself as well as the World Wide Web. In his seminal piece, ``As we may
think'', \cite{bush1945} described his vision of the Memex, explaining that
the user of the Memory Extender ``can add marginal notes and comments [\dots]
by a stylus scheme''. And \cite{bernerslee1989} described, in the original
concept note that laid the groundwork for what later became the World Wide
Web, that one ``must be able to add one's own private links to and from public
information. One must also be able to annotate links, as well as nodes,
privately.'' While Berners-Lee had this specific idea in mind already back in
1989, it took more than 20 years of work for Web Annotations to become a web
standard proper (see below).

Linguistic annotations are, procedurally, conceptually, and technically,
closely linked to markup and markup languages, especially the ones based on
XML \citep[Extensible Markup Language,][]{bray2008}, enriched, processed,
presented and queried with related formalisms such as, among others, XML
Schema, XSLT, XPath, XQuery, CSS, RDF and OWL. Through their unambiguous,
syntactic separation of annotations from the primary data, markup languages
are a natural candidate for linguistic annotations, especially those based on
XML, the most widely used meta-language for the definition of concrete markup
languages using approaches such as XML Schema or Document Type Definitions
(DTD). One of the most widely used annotation systems in Linguistics and
Digital Humanities are the TEI guidelines \citep{tei2019}, initially developed
in the late 1980s. The formalisms mentioned above were developed and
standardised by the World Wide Web Consortium (W3C), an international
non-profit organisation founded by Tim Berners-Lee in 1994 to lead the further
development of the World Wide Web's technical building blocks. Just like XML,
the W3C's effort to move from a static, document-centric to a \emph{Semantic}
Web also lead to a number of highly influential and innovative developments in
Linguistics and Computational Linguistics, especially with regard to modelling
and querying annotations \citep{rehm2007e,farrar2010,chiarcos2015}. The
interface between technical markup and linguistic annotations is examined by
\cite{metzing2010} including the interface between HTML and linguistic markup
\citep{rehm2007b}.

Most stand-alone tools for the annotation of linguistic data, often
implemented in Java, have by now vanished or, if they are still in use, target
a specific niche for which a browser-based solution has not been developed
yet. Nowadays, actual annotation work is typically carried out in the web
environment, i.\,e., in the browser, using one of the web-based annotation
tools such as, among others, Brat \citep{stenetorp2012}, WebAnno
\citep{eckartdecastilho2016}, INCEpTION \citep{klie2018} or CATMA
\citep{catma2019}. Crucially, the textual data that is annotated this way may
be web data (i.\,e., HTML documents) that was downloaded or crawled, but it is
typically \emph{not} live web data because anchoring annotations to live web
documents that can change, in a subtle or substantial way, any minute is
technically challenging.

The fairly recent W3C standard Web Annotation was developed for exactly this
purpose, i.\,e., to enable the annotation of live web data. The standard
consists of three W3C recommendations. The Web Annotation Data Model
\citep{sanderson2017a} describes the underlying annotation data model as well
as a JSON-LD serialisation. The Web Annotation Vocabulary
\citep{sanderson2017b} underpins the Data Model, and the Web Annotation
Protocol \citep{sanderson2017c} defines an HTTP API for publishing,
syndicating and distributing Web Annotations. The standard enables users to
annotate arbitrary pieces of web content in the browser, essentially creating
an additional, independent layer on top of the regular World Wide Web. Web
Annotations are \emph{the} natural mechanism to enable web users and readers,
on a general level, interactively to work with content, to include notes,
feedback and assessments, to ask the author or their peers for references or
to provide criticism. However, there are still limitations. As of now, none of
the larger browsers implement Web Annotations natively, i.\,e., content
providers need to enable Web Annotations by integrating a corresponding
JavaScript library. Another barrier for the widespread adoption of Web
Annotations are proprietary commenting systems, as used, among others, by all
major social networks who are keen on keeping all annotations (i.\,e.,
comments and other types of user-generated content) in their own respective
silos and, thus, under their own control.

Nevertheless, services such as the popular Hypothes.is tool (see below) enable
Web Annotations on any web page, but native browser support, ideally across
all platforms, is still lacking. In addition to the (still somewhat limited)
ability of handling live web data, the Web Annotation standard has multiple
advantages that make it perfectly suited for linguistic annotations. The Web
Annotation Data Model is very general and can be conceptualised as a
multi-layer Annotation Graph. Annotations are sets of connected resources,
typically an annotation \emph{body} and the \emph{target} of the
annotation. If and when the Web Annotation standard is finally available
natively in all browsers, conversations between users and content creators can
take place anywhere on the web in a standards-compliant way, where, and this
is crucial, the annotations are under the control of the users because
annotations can live separately from the documents they are pointing to --
they are reunited and re-anchored in real time.

The annotation tool developed by the non-profit organisation Hypothes.is is by
the far the most popular one. It enables taking private notes or publishing
public annotations. It can be used in collaborative groups, it provides Linked
Data connections and works with different formats including HTML, PDF and
EPUB. It is used in scholarly publishing and as a technical tool for open peer
review, in research, education and investigative journalism.\footnote{See, for
  example, the projects presented in the various events of the ``I Annotate''
  conference series, which started in 2013: \url{http://iannotate.org}.} It
can also be used for automated annotations, e.\,g., to tag Research Resource
Identifiers (RRIDs).

\pagebreak
With regard to the current state and further development of markup approaches
and technologies, XML, originally published in 1998 and, since then, in
extremely widespread use, is no longer actively maintained or developed
further within W3C. However, there is still a highly active and passionate
community interested especially in declarative markup. Discussing some of the
lessons learned during the development of XML, \cite{walsh2018} emphasise the
need for a new umbrella environment and community initiative for future work
on descriptive markup: the Markup
Declaration.\footnote{\url{https://markupdeclaration.org}}

\section{Computational Linguistics}
\label{sec:compling}

The annotation landscape, which consists, generally speaking, of tools and
formats, has had several decades to grow and to mature into an area that is
impossible to characterise in the context of a short book chapter alone. Many
colleagues provided general or specific overviews, including, among others,
\cite{bird2001}, \cite{dipper2004}, \cite{metzing2010},
\cite{stuehrenberg2012}, \cite{ide2017}, \cite{biemann2017}, \cite{stede2018},
\cite{neves2020}. In addition to a large number of all-purpose and specialised
formats \citep{ide2017c} such as, among many others, TEI, NIF, NAF, LAF, GRAF,
TIGER, STTS, FoLIA, there is a plethora of editors and tools to chose from,
such as Brat, WebAnno, Exmaralda, Praat, ELAN, ANNIS, CATMA, INCEpTION and
Prodigy as well as many others including crowd-sourced approaches.


%

Both annotation tools and also annotation formats can be described along a
number of dimensions and continuums. Annotation schemes range from
\emph{trivial} (e.\,g., marking up single tokens) to \emph{complex} (enabling
semantically deep and nuanced annotations). These often correlate with their
annotation task, from \emph{easy}, \emph{straightforward} and \emph{well
  understood} (e.\,g., annotating named entities) to \emph{hard},
\emph{challenging} and \emph{novel} (e.\,g., the annotation of actors and
events in storylines). Accordingly, simple annotation tasks, the goals of
which can be summarised and specified in concise annotation guidelines
effectively, typically result in very high inter-annotator agreement scores
while hard, ambitious and challenging tasks that may require a certain level
of expertise or training, rather result in low inter-annotator agreement
\citep{gut2004,bayerl2007,bayerl2011,snow2008,artstein2017}.
Finally, simple annotation tasks are typically carried out using general
all-purpose tools while complex annotation tasks usually require specialised
or customised tools.

\section{Artificial Intelligence}
\label{sec:ai}



Artificial Intelligence (AI) as an academic discipline was founded in the
1950s. While it consists of various subfields, by now, it is ubiquituous first
and foremost due to the recent breakthroughs made in the area of Machine
Learning (ML) using Deep Neural Networks (DNNs). These have been made possible
due to powerful supervised but also unsupervised machine learning algorithms,
fast hardware and, crucially, large amounts of data. This is why the relevance
of annotations and annotated data sets for AI at large, including
Language-Centric AI \citep{rehm2020k}, i.\,e., Computational Linguistics and
Natural Language Understanding, has increased dramatically in recent years.

Modern AI methods are data-driven. Supervised learning methods rely on very
large annotated data sets, many of which consist of primary (language) data
and secondary annotations, as defined in
Section~\ref{sec:definition}.\footnote{In Natural Language Understanding, DNNs
  are also used for language modelling, i.\,e., for generating statistical
  models out of enormous amounts of unannotated language data. These can be
  used for various classification and prediction tasks \citep{rehm2019e}.} In
fact, data curation and annotation has become so important that new business
models have emerged that revolve around the production of structured data for
customers who want to make use of supervised learning in concrete application
scenarios. Some companies employ in-house experts for the construction of data
sets while others use crowd-working approaches.\footnote{For example, Appen's
  current slogan is ``Data with a human touch: High-quality data for machine
  learning, enhanced by human interaction'' (\url{https://appen.com}).} Key
aspects of any data generation process include the annotation speed, the
quality and relevance of the annotations, and how meaningful, reliable and
representative the annotations are.


With regard to the context of AI-based applications, the line between the
construction of structured data sets on the one hand and the collection of
--~typically user-generated~-- data points on the other, is blurry, as both
can be conceptualised as annotations. In the former, language data is
annotated with regard to, for example, word senses or intents. In the latter,
actual live content is ``annotated'', for example, by liking a tweet, leaving
a five-star rating for a restaurant or commenting on a news article. All of
these activities are annotations that add metadata to existing data. Clicking
a headline to go to an article or even turning the page in an ebook can also
be and, in fact, are interpreted as annotations with regard to the underlying
primary data in question. Increasingly slower page turns in an ebook, for
example, could be interpreted by the user modelling algorithm as ``boredom''
with the current chapter and may, later on, result in automatically adjusted
book recommendations. Even the non-action of no longer reading an ebook can be
seen as an ``implicit'' annotation. In the future, for certain non-fiction
genres it will be possible to identify the chapters in which readers lose
interest and then to generate slightly different versions or paraphrases of
those chapters with the intent of not losing any readers by keeping their
engagement high. In these cases, the original human author will compete with
the machine in an A/B test, i.\,e., both variants are presented to users in a
short experimental phase, while only the statistically more effective variant
will be used in the long-term. In today's digital age, users of large online
applications must be aware of the fact that every single action or click they
perform, i.\,e., every single annotation, is recorded, associated with their
profile, and made use of by user modelling and recommender algorithms,
including advertisements.

\section{Language Technology}
\label{sec:langtech}

The applied field of Language Technology (LT) transfers theoretical results
from language-oriented research into technologies and applications that are
ready for production use. Linguistics, Computational Linguistics,
Psycholinguistics, Computer Science, AI and Cognitive Science are among the
relevant fields made use of in LT-solutions. Spell checkers, dictation
systems, translation software, search engines, report generators, expert
systems, text summarisation tools and conversational agents are typical
LT-applications.


This Section takes a brief look at potential ways how LT as well as AI can
interface with the Web Annotation technology stack (Section~\ref{sec:www}). LT
can be embedded in various phases and places of the Web Annotation workflow to
address and eventually solve a number of common challenges
\citep{presrehm2016a}. First, the web content to be enriched with annotations
can be created automatically or semi-automatically using Natural Language
Generation (NLG) approaches; in fact, this is already the case for vast
amounts of online content, including online shops, weather reports, and
articles about sport events. Second, the web content can be automatically
analysed and then annotated using LT, for example, for the purpose of
generating an abstract of a longer article using automated text summarisation
and then presenting the article to users in the form of an annotation. Third,
the content of the actual annotations, potentially made by many different
users, can be analysed using LT, for example, for the purpose of mining the
feedback of the users or readers for sentiments and opinions towards the
primary content, which may be a product description, a news article on a
breaking event or a discussion of a topic of high social relevance. In that
regard, web annotations are also --~just like blogs, online videos, online
photos~-- User-Generated Content (UGC). Currently, with individual silos
containing UGC, it is complex, challenging and costly to perform Social Media
Analytics and Opinion Mining at scale due to the various formats and
heterogeneous sources. A centralised approach based on Web Annotation would
simplify such text mining approaches significantly, also enabling a much
broader and more varied analysis of opinions regarding, among others,
commercial products, societal challenges, political trends and misinformation
campaigns \citep{rehm2017o,rehm2017j,rehm2018c,rehm2017m}.

The Web Annotation standard is based on the notion of stand-off annotation,
i.\,e., the annotations are not embedded inline within the actual primary data
in the form of, e.\,g., XML elements, but stored indepedently from the primary
data. This approach enables overlapping annotations, i.\,e., stand-off
annotations do not have to adhere to the rather strict requirements regarding
the tree structure imposed by the XML standard. Instead, stand-off annotations
make use of a pointing or linking mechanism so that an annotation is anchored
to or linked to a certain sequence of primary data. This (important) advantage
comes with a computational cost, though, because each stand-off annotation
needs to be explicitly anchored at processing time. In our recent and current
research projects\footnote{DKT (\url{http://digitale-kuratierung.de})
  \citep{rehm2016j}, QURATOR (\url{https://qurator.ai}) \citep{rehm2020c} and
  LYNX (\url{http://lynx-project.eu}) \citep{rehm2019c}.} we use a similar
approach, the NLP Interchange Format \citep[NIF, see][]{hellmann2013}. NIF was
developed especially for LT applications and is based on the Linked Data
paradigm, i.\,e., RDF and OWL.

Between the development phase and the deployment phase of an LT-based
solution, annotation formats can also be mixed. For example, in LYNX, all
processing solutions make use of NIF \citep{rehm2019c} but during the
development and training phase of the German Legal NER model we used the CONLL
format which is a simple, tab-seperated value, i.\,e., non-XML-based inline
annotation format \citep{rehm2019d,leitner2020b}.

\section{Open Science}
\label{sec:openscience}

The umbrella term Open Science denotes the movement to make scientific
research, data and dissemination accessible to interested stakeholders. It
includes a multitude of different aspects, e.\,g., publishing open research,
pushing for Open Access (instead of closed) and encouraging researchers of all
fields to publish not only their results but also their data for easier
verification and reproducibility. Open Science is becoming more and more
popular and is, crucially, relevant to the broader topic of annotations. If we
examine the taxonomy\footnote{See
  \url{https://www.fosteropenscience.eu/foster-taxonomy/open-science}.}
produced by the EU project FOSTER to describe the different aspects of Open
Science, these connections become immediately apparent: Open Science advocates
for Open Data, which should not only be open but also annotated using
standards, made available using platforms that are accessible (e.\,g., Linked
Data) and described with metadata and semantics including well defined
categories and taxonomies.

One of the key goals of promoting Open Research Data is to enable data re-use
and, thus, Open Reproducible Research that also includes Open Science
Workflows, often made possible by distributing Open Source software and
specifying the workflows used to arrive at the results published in a
scientific article. Annotations, the meaning and semantics of which are
clearly documented, ideally using international standards, are the glue
between the software components that produce the annotations, annotated open
research data, annotation guidelines, research data repositories, query
mechanisms and scientific publications.

With the ever growing and maturing technology infrastructure for
data-intensive research, Open Science will soon become the norm, including the
use of sustainable repositories for making available research data clearly
described and annotated using standardised, best-practice approaches, linked
to other sets of research data, fostering the re-use of the data in the
context of new research questions. The FAIR Data Principles emphasise, in
their procedural order, four main aspects of research data, which should be
made findable, accessible, interoperable and re-usable
\citep{fair2016}.\footnote{See \url{https://www.go-fair.org} for more detailed
  descriptions of the principles.} Most of the FAIR principles refer to
metadata, which can, especially if they relate to primary data, also be
conceptualised as annotations. The relevant principles are the following ones:

\medskip
\begin{compactdesc}\small
\item[F2] Data are described with rich metadata.
\item[F3] Metadata clearly and explicitly include the identifier of the data
  they describe.
\item[A1] (Meta)data are retrievable by their identifier using a standardised
  communications protocol.
\item[A2] Metadata are accessible, even when the data are no longer available.
\item[I1] (Meta)data use a formal, accessible, shared, and broadly applicable
  language for knowledge representation.
\item[I2] (Meta)data use vocabularies that follow FAIR principles.
\item[I3] (Meta)data include qualified references to other (meta)data.
\item[R1] (Meta)data are richly described with a plurality of accurate and
  relevant attributes.
\item[R1.2] (Meta)data are associated with detailed provenance.
\item[R1.3] (Meta)data meet domain-relevant community standards.
\end{compactdesc}
\medskip

As can be seen, the FAIR principles --~and also Open Science in general~--
recommend, at their core, the use of standards for the purpose of enabling or
enhancing, as much as possible, the findability, accessibility,
interoperability and reusability of research data \citep[see][for a practical
example]{labropoulou2020l}. While these recommendations are important and,
thus, to be supported, it is also worth noting that especially basic research
is about trying and inventing \emph{new} things, i.\,e., things that have,
almost by definition, \emph{not} been standardised yet. This contradicts, on a
fundamental level, with the recommendation of using standards as the consensus
reached within a specific research community to represent, for example,
temporal expressions in natural language text. The contradiction can be
resolved, though, if the recommendation is relaxed to the use of established
tools and best practice approaches as well as the modification and extension
of standards. The crucial aspect is to document the semantics of the
annotation scheme used in a corpus or data set. If an established,
standardised approach does not work for an emerging piece of research, a new
approach needs to be created or an established approach modified.

It is safe to predict that Open Science will be transforming research in the
next years, making it more sustainable, more visible and more
transparent. Several disciplines have already been following Open Science-like
approaches for quite a while. On a larger scale, though, Open Science will
only be fully possible with substantially improved digital infrastructures.
Notable initiatives are the European Open Science Cloud
(EOSC)\footnote{\url{https://ec.europa.eu/research/openscience/index.cfm?pg=open-science-cloud}}
and the Nationale Forschungsdateninfrastruktur
(NFDI)\footnote{\url{https://www.dfg.de/foerderung/programme/nfdi/}} in
Germany. Additionally, we can predict that, soon, robust and large-scale
services for the annotation of documents will be provided, starting with
scientific publications, for which it will be possible to annotate and, thus,
explicitly represent, using standardised metadata schemas and ontologies,
their methods used or expanded upon, evaluation approaches, data sets as well
as findings and contributions -- this structured set of semantic information
associated with one research article, as the atomic unit of scientific
publication, will be contextualised in larger knowledge graphs which will
capture the research output of entire scientific fields, including
annotations. Several larger scientific publishing houses are already now
developing corresponding digital infrastructures to capture the results they
publish. At the same time, the Open Research Knowledge Graph (ORKG) initiative
promotes the vision of moving scholarly publishing from a coarse-grained,
predominantly \emph{document-based} to a \emph{knowledge-based} approach by,
first, automatically identifying and extracting and, second, representing and
expressing scientific knowledge through semantically rich, interlinked graphs
\citep{orkg2019}.\footnote{\url{https://www.orkg.org}} In a third step, the
knowledge contained in the ORKG can be used, for example, to compare the
approaches followed in different scientific papers on the same research
question.

\section{Dimensions of Annotations} 
\label{sec:dimensions}

\begin{figure}[htbp]
  \centering
  \includegraphics[width=.9\textwidth]{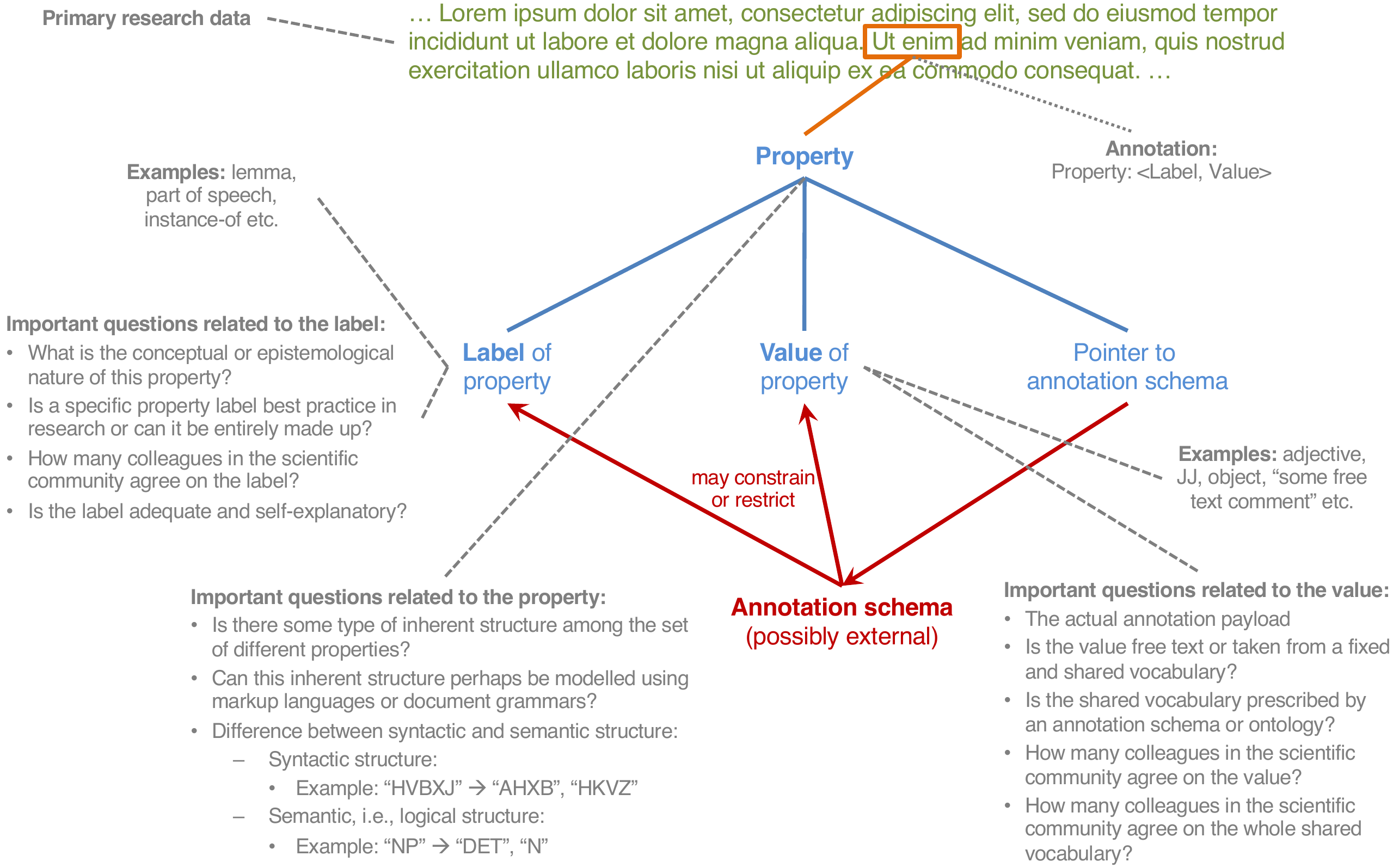}  
  \caption{General aspects and dimensions of annotations}
  \label{fig:property}
\end{figure}

The process of adding annotations to a set of primary research data can be
conceptualised as the insertion of secondary research data (see
Section~\ref{sec:definition}). The secondary data added to the primary data
typically refers to one or more (often interconnected) properties of the
primary data that are explicitly marked using syntactically identifiable
methods. Figure~\ref{fig:property} shows the general aspects and dimensions
involved in an annotation in more detail; \cite{ide2001} provide a similar but
more technical view focused upon syntactic annotations.

An annotation explicitly describes a \emph{property} of a piece of primary
data using a tuple that consists of the \emph{label} of the property in
question (e.\,g., ``part of speech'') and a corresponding \emph{value}
(e.\,g., ``adjective''). An annotation can also include a pointer to an
abstract, internally or externally represented annotation scheme that,
typically, specifies the semantics of all possible annotations. This
annotation scheme, in turn, can be used to constrain or to restrict specific
annotations, i.\,e., the <label, value> pair that makes up an annotation.

Especially when designing a new or modifying an existing annotation scheme to
address a specific research experiment, several relevant questions need to be
taken into account, some of which are included in
Figure~\ref{fig:property}. These question pertain, among others, to the
conceptual or epistemological nature of the specific label of an annotation:
on the one hand, this label can denote a concept that has been established in
a scientific field for decades or it can refer to a fairly recent aspect,
phenomenon or finding, for which an established term in the respective
scientific community does not exist yet. Another aspect relates to the set of
properties that are being described with the help of an annotation scheme: are
these isolated properties without any inherent structure that governs the
sequence or distribution of their instantiations (e.\,g., different types of
named entities) or does some kind of linguistic or syntactic structure exist
on top of the different annotations? If the latter is the case, can this
structure be explicitly modelled, for example, using mechanisms built into XML
DTD or XML Schema-based document grammars \citep{maler1996,megginson1998}?
Can, maybe as an additional mechanism on top of the document grammar, an
ontology be used to describe higher-level semantic concepts?

The various notions hinted at in Figure~\ref{fig:property} lead us to a more
abstract aspect of annotations: just like primary research data, annotations
have various properties themselves. Depending on the research question and
overall use case, it may be important or even necessary to explicitly
represent these properties, i.\,e., to annotate annotations. Among this set of
properties are the following: \emph{annotator} of the annotation (i.\,e., was
it created by a human expert or by an automatic process?), \emph{annotation
  layer} (i.\,e., does the annotation refer to the ``document structure'',
``layout'', ``syntax'', ``semantics'', ``information structure'' etc.?),
\emph{confidence value} (i.\,e., how confident is the human annotator or
automated process that an annotation is correct?), \emph{timestamp} (i.\,e.,
when the annotation was added), \emph{style} (i.\,e., how an annotation is
rendered in a certain system) and \emph{application scenario} (i.\,e., is the
annotation primarily meant for human or machine consumption?). It is important
to note that more structure can be explicitly added even on top of these
annotations, especially with regard to the relationship and interdependence of
the various annotation layers.

Instantiated sets of annotations can be described along various axes and
dimensions, some of which are rather vague while others are more concrete.

\begin{itemize}
\item \emph{Annotator:} The actual source or origin of annotations included in
  a data set, for example, one or more automated components, human experts,
  human laypersons, crowd workers etc. This dimension also refers to the
  \emph{methodology} followed for including the annotations into the primary
  data.
\item \emph{Semantics:} The semantics of the annotations, i.\,e., the nature
  of the properties explicitly and formally described through the annotations,
  e.\,g., linguistic concepts or aspects relating to document structure,
  rhetorical structure, genre, style, terminology etc. This dimension is
  connected to the \emph{annotation scheme} used, which could be an
  experimental scheme developed, e.\,g., in a research project for a novel
  purpose, or one of the well known annotation schemes and standards that have
  been in use for decades, e.\,g., TEI.
\item \emph{Layers:} The nature and interconnectedness of the different
  annotation layers if an annotated data set contains multiple layers.
\item \emph{Guidelines:} A crucial question with regard to annotation projects
  primarily carried out by humans, relates to the presence of annotation
  guidelines, especially with regard to the specification of concrete examples
  and exceptions, i.\,e., which concepts to annotate how in a specific
  context.
\item \emph{Research question or application use case:} An annotated data set
  is typically associated either with an underlying research question that has
  motivated the construction of a data set or with a concrete annotation
  pipeline (i.\,e., application use case) that was used to annotate the
  primary data.
\item \emph{Complexity:} This dimension refers to the notion that some
  annotations are more complex than others, it is closely related to several
  other dimensions.
\item \emph{Evaluation:} Most annotated data sets have been evaluated in some
  way, e.\,g., with regard to the inter-annotator agreement (if the primary
  data was annotated by multiple annotators).
\end{itemize}

Space restrictions prevent us from describing all dimensions in more detail,
which is why we concentrate on \emph{Complexity}
(Section~\ref{sec:complexity}) and \emph{Evaluation}
(Section~\ref{sec:evaluation}).

\subsection{Complexity of Annotations}
\label{sec:complexity}


In Computational Linguistics and also in the wider Digital Humanities area,
several fairly detailed annotation schemes and markup languages have been
developed for the annotation of textual data in the last 30 years. The TEI
guidelines are probably the most extensive ones -- the PDF version of the TEI
P5 guidelines \citep{tei2019} has a length of almost 2000 pages, in which
hundreds of XML elements and attributes, grouped into various modules, are
described. In stark contrast, the annotation schemes used in many current data
sets, especially for large-scale, data-driven AI approaches that rely on vast
amounts of training data, are quite shallow and highly generalised. Machine
learning approaches perform best with large amounts of training data; it is
beneficial for the performance of the resulting models and classifiers if the
number of unique class labels is rather small and the number of different
examples per class label rather high. Especially for environments in which
such AI-based classifiers are used in production, the corresponding data sets
are often created by professional annotation teams or companies (see
Section~\ref{sec:ai}). In these scenarios and use cases it is not feasible to
annotate data sets with complex annotation schemes. 

It is an interesting question for future research if the difference in
complexity or the ``level of sophistication'' of different annotation schemes
--~from a simple set of a few labels to highly complex markup languages like
TEI P5~-- can be measured or formally described. To the best of the author's
knowledge, there has not been any work on this topic so far. Many different
data points and statistics about an annotation scheme could be exploited for
this purpose, e.\,g., the number of property labels (i.\,e., XML tags), the
number of meta properties (e.\,g., XML attributes), the number of free text
and predefined values, the presence of inherent structure including nesting
levels etc. These, and other, statistics could be included in a formula that
captures the complexity of an annotation scheme; it could also be used,
together with data such as token/annotation ratio, to model the complexity of
the annotations contained in a concrete data set.

\subsection{Evaluation of Annotations}
\label{sec:evaluation}

The evaluation of annotations is a crucial dimension of formally describing a
data set or corpus, especially when it was created for the purpose of training
a practical tool and also when an emerging annotation scheme was used. In that
regard, two different aspects can be evaluated that are intricately
interrelated: the annotation scheme itself and concrete annotations.

The evaluation of the validity of an abstract, possibly emerging, annotation
scheme is typically an iterative process \citep{dickinson2017,artstein2017}:
first, an initial version of the annotation scheme is applied to a small and,
ideally, representative data set to examine if it is practical and balanced
concerning its ability to annotate all the characteristics and phenomena it is
supposed to be able to mark up explicitly. An overarching aspect that should
be taken into account when developing and iteratively evaluating an annotation
scheme relates to the question if it models scientific consensus. These
initial tests are, later on, repeated with more mature versions of the
annotation scheme until all requirements, prescribed by the respective
research question, are met. As the two go hand in hand, these initial
evaluations typically concern not only the annotation scheme but also the
annotation guidelines as well as their applicability using a specific
annotation tool. Important questions regarding the annotation guidelines
relate to their length, coverage, examples, and exceptions as well as how long
it usually takes to train annotators so that they can perform an annotation
task.

The result of an annotation task or process can also be evaluated, both
qualitatively and quantitatively. In the context of this chapter, the typical
approach is to compare multiple annotations of the same primary data, created
by multiple annotators, and to compare their inter-annotator agreement,
i.\,e., how well do the various annotators agree when comparing their
respective annotations. Multiple approaches to calculate inter-annotator
agreement exist \citep{gut2004,bayerl2007,bayerl2011}. This analysis is
crucial for data and experiment-related aspects such as replicability and
reproducibility and for measuring the consensus among the annotators,
especially for complex annotation tasks or emerging annotation formats. A
variation of measuring inter-annotator agreement can be described as
``intra-annotator agreement'', i.\,e., the same annotator is asked to perform
the same annotation task multiple times but under different conditions or
several days or weeks apart. This approach can also be used to identify
weaknesses in emerging annotation schemes or guidelines.

\section{Summary and Conclusions}
\label{sec:conclusions}


This article presents various observations on annotations. It approaches the
topic from multiple angles including Hypertext, Computational Linguistics and
Language Technology, Artificial Intelligence and Open Science. Annotations can
be examined along different dimensions. In terms of complexity, they can range
from trivial to highly sophisticated, in terms of maturity from experimental
to standardised. Annotations can be annotated themselves using more abstract
annotations. Primary research data such as, e.\,g., text documents can be
annotated on different layers concurrently (e.\,g., general segmentation
including text structure, coherence relations, syntax), which are independent
but can be exploited using multi-layer querying. Standards guarantee
interoperability and reusability of data sets, which is especially crucial in
terms of Open Science.

The chapter concludes with four final observations, formulated as research
questions or rather provocative remarks on the current state of the field.

\textbf{Do standards hold back innovative annotation research?} Standard
annotation schemes represent the condensed consensus gathered within a wider
research community regarding certain phenomena. This class of standardised
formats is crucial for interoperability and reproducibility. However, one
aspect that is often neglected concerns the fundamental nature of research
itself, which is about finding, creating and inventing \emph{new} things, new
pieces of knowledge, new insights, including new ways of annotating language
data. Especially taking into account those annotation schemes that are, both
conceptually and also technically, highly similar, it is worth emphasising
that new breakthroughs require new approaches. Focusing on standards too much
may hold back research.


\textbf{Can we concentrate on annotating live web data instead of dead web
  data?} Primary research data is nowadays typically annotated within a
web-based environment, i.\,e., using a dynamic web application that visualises
both the primary and the secondary research data in a browser. Very often,
said primary data is, in fact, web data, i.\,e., text or multimedia data that
was either crawled or collected using other means from the World Wide
Web. Crawling and archiving live web data decouples the documents from their
natural habitat, which essentially results in frozen snapshots of these
documents. While this approach has been best practice in Computational
Linguistics almost since the beginning of the World Wide Web, it would be much
more interesting to treat the \emph{live} World Wide Web as a corpus. Given
that the web technology stack even includes its own annotation approach (Web
Annotation, see Section~\ref{sec:www}), we should attempt to treat the whole,
live World Wide Web as a giant corpus by parsing the whole web and by adding
linguistic information using the Web Annotation approach, which can then be
queried for linguistic analyses or for training machine learning models
\citep{rehm2017j,rehm2018c}. To that end, larger collections of web-native
Language Technology services \citep{elg2020,rehm2020c} could be used in
high-performance infrastructures \citep{rehm2020n}.

\textbf{Is it possible to design a machine-readable packaging format for
  describing annotations?} Annotations have different dimensions along which
they can be described (Section~\ref{sec:dimensions}). It would be a highly
interesting question to examine if it is possible to design a compact,
machine-readable packaging format for describing annotation projects including
the annotations themselves as well as the overall approach, main formal
aspects of the annotation scheme (including its complexity) and the concrete
annotations. This is a relevant and important question from the point of view
of Open Science (and more transparent as well as reproducible and
interoperable science). The question also relates to machine learning,
language resources and emerging AI and LT platforms. Soon, these will be able
to import a data set and use a machine learning toolkit automatically to train
a new model \citep{rehm2020n}. In order for this to work fully automatically,
we need metadata schemes to describe annotated data sets including formal
aspects such as their annotation schemes and involved dimensions.

\textbf{Is the field ignoring decades of valuable annotation science
  research?} Since the emergence of the first large corpora and the
statistical turn in the early 1990s, Computational Linguistics has produced a
plethora of results and insights regarding the annotation of language
resources -- so much so that \cite{ide2007b} even speaks of ``annotation
science''. In the last five years, neural approaches have turned out to be
very popular in Language Technology, outperforming essentially all of the
previous methods. Generally speaking, neural technologies require very large
data sets for training models. Corresponding applications are often
generalised as classification tasks that are based on large data sets that
were annotated with only few labels. In many cases, both the classification
tasks and also the sets of labels or annotations must be described as rather
simplistic, often focusing upon incremental research challenges. At the same
time, many of the recent language resources were annotated on a rather shallow
level, with only a few highly generalised and abstract labels, often using
crowd-workers who are only able to produce large amounts of consistent and
high quality annotations if the annotation task is rather simple and does not
require expert linguistic knowledge or in-depth training \citep[][call these
``microtasks'']{poesio2017a}. In short, since the neural turn in
approx.~2014/2015 we can observe a trend towards \emph{simply more and more}
annotations with increasing quantity while ignoring complexity and structure,
and also a trend towards \emph{more and more simple} annotations that are
cheaper to produce and easier to generalise from. Has annotation science
perhaps become obsolete? Have the lessons and insights learned in the last
30~years become irrelevant, given today's popularity and power of neural
approaches for processing and, perhaps, finally, understanding language?

\section*{Acknowledgement}

This chapter is based on a presentation given at the conference
\emph{Annotation in Scholarly Editions and Research: Function --
  Differentiation -- Systematization}, held at the University of Wuppertal,
Germany, on 20-22 February 2019. The author would like to thank the
organisers, Julia Nantke and Frederik Schlupkothen, for the invitation and
especially for their patience. Peter Bourgonje and Karolina Victoria Zaczynska
provided comments on an early draft of this article for which the author is
grateful. Work on this chapter was partially supported by the projects ELG (EU
Horizon 2020, no.~825627), LYNX (EU Horizon 2020, no.~780602) and QURATOR
(BMBF, no.~03WKDA1A).

\bibliographystyle{plainnat}
\bibliography{rehm.bib,literatur.bib,hola.bib}

\end{document}